\documentclass[10pt,twocolumn,letterpaper]{article}

\usepackage{cvpr}              

\usepackage{graphicx}
\graphicspath{{graphs/}}
\usepackage{amsmath}
\usepackage{amssymb}
\usepackage{booktabs}

\usepackage{amsfonts}

\usepackage{siunitx}
\usepackage{comment}
\usepackage{caption}
\usepackage{subcaption}
\usepackage{multirow}

\usepackage[pagebackref,breaklinks,colorlinks]{hyperref}

\usepackage[capitalize]{cleveref}
\crefname{section}{Sec.}{Secs.}
\Crefname{section}{Section}{Sections}
\Crefname{table}{Table}{Tables}
\crefname{table}{Tab.}{Tabs.}

\def\cvprPaperID{4473} 
\def\confName{CVPR}
\def\confYear{2022}

\begin{document}

\title{Self-Supervised Bulk Motion Artifact Removal in Optical Coherence Tomography Angiography}
\author{Jiaxiang Ren\textsuperscript{1}\quad Kicheon Park\textsuperscript{2}\quad Yingtian Pan\textsuperscript{2}\quad Haibin Ling\textsuperscript{1}\thanks{Corresponding author.}\\
\textsuperscript{1}Department of Computer Science\quad  \textsuperscript{2}Department of Biomedical Engineering\\
Stony Brook University\\
{\tt\small \{jiaxren,hling\}@cs.stonybrook.edu\quad \{ki.park,yingtian.pan\}@stonybrook.edu}}
\maketitle

\begin{abstract}

Optical coherence tomography angiography (OCTA) is an important imaging modality in many bioengineering tasks. The image quality of OCTA, however, is often degraded by Bulk Motion Artifacts (BMA), which are due to micromotion of subjects and typically appear as bright stripes surrounded by blurred areas. 
State-of-the-art methods usually treat BMA removal as a learning-based image inpainting problem, but require numerous training samples with nontrivial annotation. In addition, these methods discard the rich structural and appearance information carried in the BMA stripe region. 
To address these issues, in this paper we propose a self-supervised content-aware BMA removal model. First, the gradient-based structural information and appearance feature are extracted from the BMA area and injected into the model to capture more connectivity. Second, with easily collected defective masks, the model is trained in a self-supervised manner, in which only the clear areas are used for training while the BMA areas for inference. With the structural information and appearance feature from noisy image as references, our model can remove larger BMA and produce better visualizing result. In addition, only 2D images with defective masks are involved, hence improving the efficiency of our method. Experiments on OCTA of mouse cortex demonstrate that our model can remove most BMA with extremely large sizes and inconsistent intensities while previous methods fail.

\end{abstract}

\section{Introduction}
\label{sec:intro}

\begin{figure}[t]
  \centering
   \includegraphics[width=1.0\linewidth]{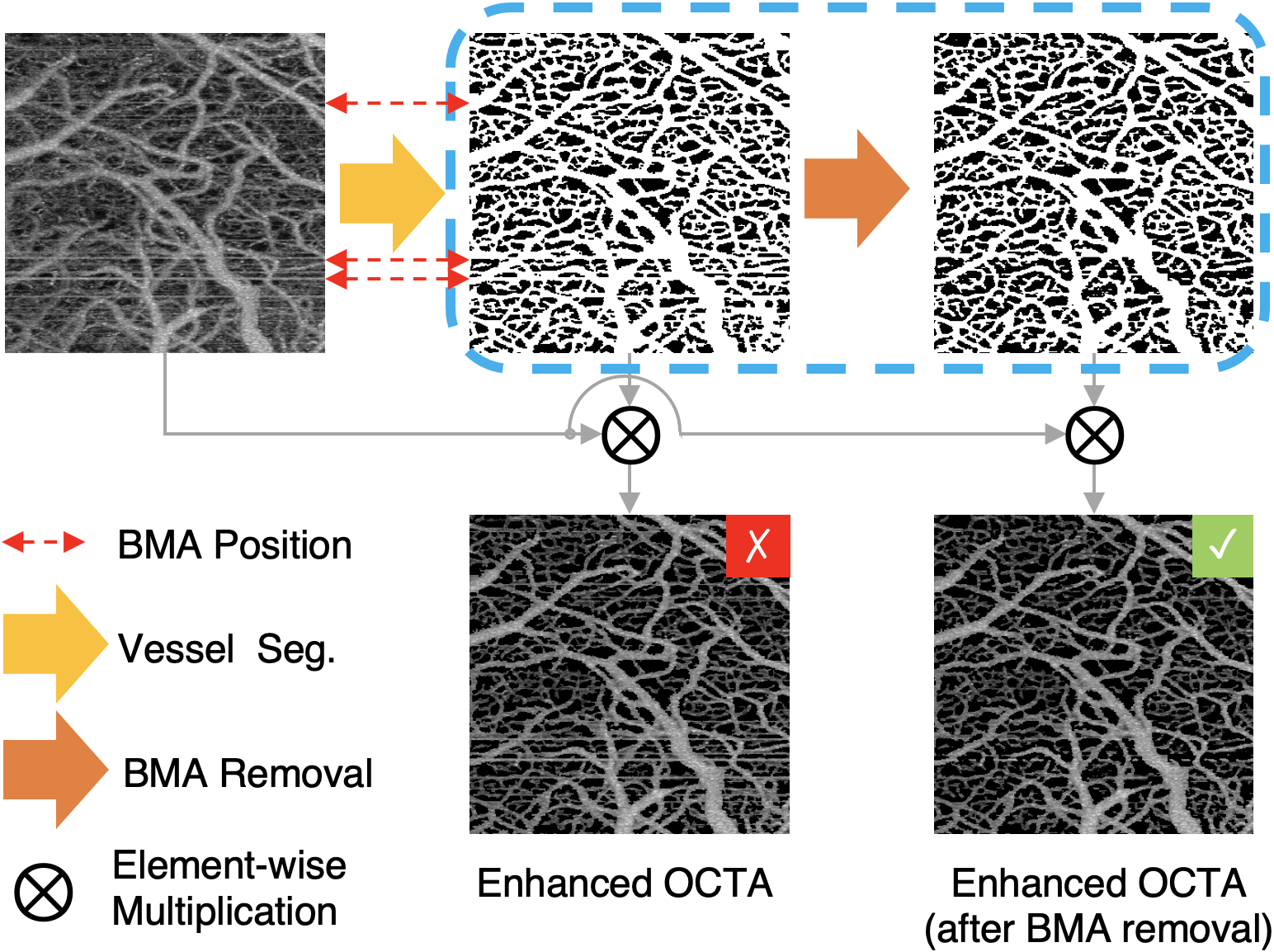}
   \caption{Illustration of OCTA affected by Bulk Motion Artifacts (BMA). Our work focuses on the BMA removal task in vessel mask (blue rectangle in dashed line). The enhanced OCTA after BMA removal has better visualization quality.}
   \label{fig:illu}
\end{figure}

As a fast and non-invasive optical imaging technology with high spatiotemporal resolution, optical coherence tomography (OCT) is an emerging medical imaging modality not only for laboratory research but also for clinical applications~\cite{huang1991optical}. Optical coherence tomography angiography (OCTA) has been reported to effectively diagnose and assess many retinal conditions, such as diabetic retinopathy~\cite{samara2017quantification}, macular degeneration~\cite{waheed2016optical} and choroidal neovascular membrane~\cite{querques2017optical}. Not only in ophthalmology, OCT has also been employed to visualize the prognosis of tumor vasculature for a better understanding of microenvironment~\cite{vakoc2009three}. Benefitting from high spatiotemporal resolution and non-invasive features, OCT has been utilized to study neurovascular changes and brain function \textit{in~vivo}.

A typical OCTA image of mouse cortex is shown in \cref{fig:illu}. Previous approach~\cite{li2017automated} employed image enhancement to remove background noises for better qualitative and quantitative analysis. Specifically, a vessel mask was generated to highlight vasculature while suppressing the background noise. Unfortunately, this approach failed to remove Bulk Motion Artifact (BMA), which was caused by micromotion of the object, as highlighted with double arrows in \cref{fig:illu}.

BMA is one of the most frequent artifacts affecting many medical imaging modalities (CT, MRI and OCT), which seriously degrades image quality and affects quantitative analysis. Over decades of research, numerous models have been developed to remove BMA, but effective solutions for all circumstances remain missing. 
In OCTA, BMA causes a severely blurred image and usually appears as horizontal stripes of different widths with high-intensity in the center and low contrast in the surrounding area. As illustrated in \cref{fig:illu}, the horizontal stripes across OCTA are the areas affected by BMA. In the following sections, we use BMA and stripe (artifacts) interchangeably. 

Previous approaches remove BMA with image registration or supervised denoising models, requiring duplicate scans or numerous training data with annotations. 
The current state-of-the-art model~\cite{li2021deep} inpaints the BMA-affected areas to generate a clear vasculature mask for enhancement. However, the inpainting model abandons the information in noisy areas so has limited capability to fill large gaps. Besides, lots of images with annotations are needed for training. We observed that, even if the intensity value of BMA is far stronger than normal area, there were still textures within BMA. Furthermore, BMA only affects vertical gradient rather than horizontal ones due to the OCTA imaging process. These features could serve as structural references of vasculature in noisy BMA area. However, there are three main challenges for BMA removal: (1) information within the motion artifacts is too noisy; (2) how to inject structure information from the BMA area into the recovery framework; and (3) OCTA data are limited and the pixel-level annotations are difficult to acquire.

In this paper, we propose a self-supervised BMA removal method to solve the issues above. First, the gradient-based structural information and appearance feature are extracted from BMA and injected into the model to capture more connectivity. Specifically, we keep the BMA-affected area, even the noisy one, as an input channel instead of discarding it directly. Furthermore, based on the observation that BMA mainly affects vertical gradients rather than horizontal ones, we introduce horizontal gradients to make our model aware of potential structures under BMA. Adding the gradient map makes the training stabler and convergence faster. With structural and appearance information and easily collected defective masks, we train our model in a self-supervised manner that the clear areas are used for training while the BMA areas are only for inference.

With the rich structural and appearance information carried in the BMA stripe region as reference, our model can remove large BMA and handle the background noise at the same time. Our BMA removal model is end-to-end and only requires 2D image with defective mask, which is more practical. Compared with other context-guided denoising models, our content-aware model can learn guidance directly from noisy images and is thus more robust. Experiments on a mouse cortex OCTA dataset demonstrate that our model can remove most BMA under different intensities and extreme area sizes while existing methods fail. The specific contributions of our paper are: 
\begin{itemize}
  \item We propose a new structural denoising model that can leverage the rich structural and appearance information carried in the BMA stripe region. 
  \item Our model is trained in a self-supervised manner, and training does not involve manual annotation.
  \item For image enhancement, our method can alleviate BMA and background noises simultaneously, and thus significantly improve image quality.
\end{itemize}

\section{Related Work}
\label{sec:relatedwork}
In this section, we will first discuss OCTA and BMA. Then, we will briefly introduce some general denoising methods followed by related OCT denoising methods.

\begin{figure*}[t]
  \centering
   \includegraphics[width=1\linewidth]{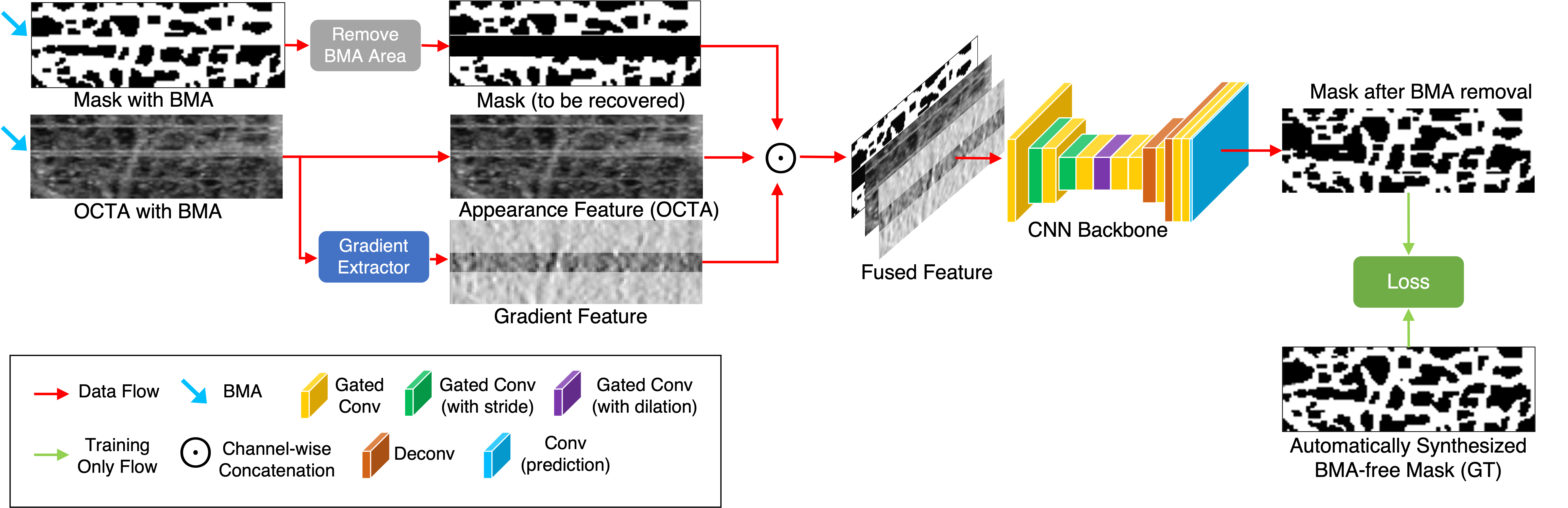}
   \caption{Framework of the proposed Content-Aware BMA Removal model (CABR).}
   \label{fig:framework}
\end{figure*}

\subsection{OCTA and BMA in Awake Animal Study}
\label{subsec:bma}
OCT is a high-speed 3D imaging technique with micron-level resolution. Previous methods employ gradient-based filters~\cite{frangi1998multiscale,law2008three,law2010oriented} to generate vessel masks to highlight vasculature for better analysis. However, the image reconstruction is sensitive to micromotion caused by heartbeat, respiration, and mechanical vibration. Such motions lead to misalignment and phase noise which cause BMA. Severe BMA usually results in extremely high-intensity noise. Previous gradient-based approaches~\cite{frangi1998multiscale,law2010oriented} can not handle such artifacts because BMA has similar gradient information with vessel branches. Moreover, long-lasting BMA could result in a broad and blurred area, making it hard to correct. Last but not least, BMA is often accompanied by background noise, which makes it even harder to find vessels within the affected area. The interleaved background noise and BMA could lead to poor visualization and erroneous quantification.

\subsection{General Denoising Method}
BM3D~\cite{dabov2007image} is a classic denoising model based on sparse representation. Inspired by the observation that natural images consist of repeated textures, BM3D groups related image patches and filters them together to shrink noise. It handles Gaussian noise well but fails to remove structural noise. After convolutional neural network (CNN) is firstly applied for the denoising task in~\cite{jain2008natural}, lots of works have been proposed to keep updating the state-of-the-art performance~\cite{chen2016trainable,zhang2018ffdnet,weigert2018content,zhou2020awgn}. Recently, Wu~\etal~\cite{wu2021contrastive} propose a single image dehazing method based on contrastive learning and achieves state-of-the-art performance on both synthetic and real-world datasets. Still, this method is trained in a supervised manner. It is not the case for some medical image modalities without ground truth.

More and more unsupervised learning based denoising methods~\cite{krull2019noise2void,lehtinen2018noise2noise,fadnavis2020patch2self} have been proposed and achieved promising performance. However, these approaches are not intended to remove structural noise. Broaddus~\etal~\cite{broaddus2020removing} employ structural kernels to remove spatially correlated noise in fluorescence microscope images. However, the performance of this method depends on the selection of blind mask, which is not fully automatic and needs prior knowledge for noise distribution.

\subsection{OCT Denoising Method}
Most traditional OCTA denoising methods~\cite{wei2017automatic,camino2016evaluation}  are based on duplicate sampling, in which BMA-affected area is replaced by BMA-free area at the same location but different sampling period. The downside of such approaches is the considerably long sampling time, which reduces the practicability for researches and clinical diagnoses. Devalla~\etal~\cite{devalla2019deep} propose a deep learning based OCT denoising model for retinal images. It is relatively easier to denoise OCT images of the retina than other tissues, such as mouse bladder and cortex, because of shallower imaging depth, less background noise, and BMA. Li~\etal~\cite{li2018automated} remove BMA in an enhancement approach. Firstly, this method applies optimally oriented flux (OOF) ~\cite{law2008three,law2010oriented,li2017automated} to generate vessel mask with defects in BMA areas and then employs tensor voting~\cite{medioni2000tensor} to restore the defective mask. Finally, an OCTA image is enhanced with the corrected mask for BMA removal. However, this method may fail to remove moderate and severe BMA.

The current state-of-the-art model~\cite{li2021deep} removes BMA with the combination of a BMA detection module, a vessel segmentation module and a mask inpainting module~\cite{yu2019free}. The BMA detection module predicts BMA affected rows in OCTA. The segmentation module takes OCTA with BMA areas removed as input and produces a corrupted mask. After that, the mask inpainting module takes the probability map of the segmentation module as input and fills the missing mask. Manually annotated masks are used to train the segmentation module and inpainting module. The whole pipeline~\cite{li2021deep} is relatively ad-hoc and the performance depends on every step. Furthermore, numerous 2D and 3D training data are involved in training. Last but not least, this model learns only from surrounding BMA-free areas while ignoring the structural information within noisy data. So this context-based model often fails when dealing with severe BMA. 

\begin{figure}[t]
  \centering
   \includegraphics[width=0.95\linewidth]{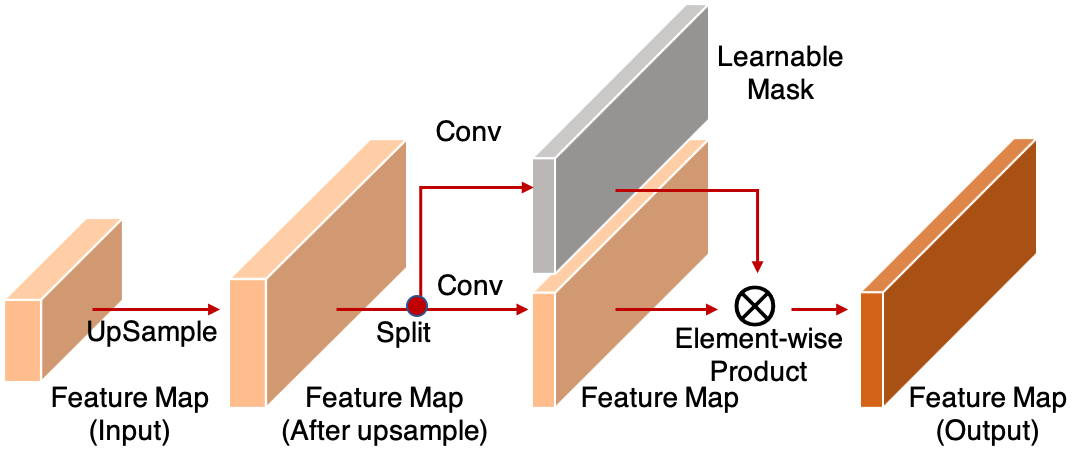}
   \caption{Deconvolution layer with GatedConv.}
   \label{fig:deconv}
\end{figure}

\section{Method}
\label{sec:method}
In this section, we will start with problem formulation and challenges. Then, to deal with the context learning issue, we will discussing two sets of information injection and the proposed Content-Aware BMA Removal Model (CABR). Finally, we will illustrate a self-supervised training strategy to solve data insufficiency.

\subsection{Problem Formulation}
We first define the BMA-affected OCTA image as $\boldsymbol{I}\in\mathbb{R}^{H\times W}$, the defective vasculature mask $\boldsymbol{M}\in\{0,1\}^{H\times W}$, the Gradient Statistics $\boldsymbol{G}\in\mathbb{R}^{H\times W}$ (See \cref{subsec:cabr}), and the corresponding row label $\boldsymbol{l}\in\{0,1\}^{H}$, where $W$ and $H$ are image width and height, respectively. $\boldsymbol{l}_i=0$ means the $i$-th row of $\boldsymbol{I}$ is clear while $\boldsymbol{l}_i=1$ for BMA. For each $\boldsymbol{I}$, the corresponding mask $\boldsymbol{M}$ is automatically generated with OOF~\cite{law2008three,law2010oriented}. The BMA affected region can be discerned easily and the row label $\boldsymbol{l}$ indicates which rows in image are affected by BMA. The proposed BMA removal framework takes $(\boldsymbol{I}, \boldsymbol{M}, \boldsymbol{G}, \boldsymbol{l})$ as input and outputs the vasculature prediction $\boldsymbol{M}'$ with BMA removed:
\begin{equation}
\label{eq:pred}
    \boldsymbol{M}' = f(\boldsymbol{M},\boldsymbol{I},\boldsymbol{G},\boldsymbol{l}).
\end{equation}


The main issue of previous context-guided inpainting models is that little information from noisy area is automatically learned. Therefore these surrounding-based generative models produce plausible but unconvincing results, especially for large missing areas. But the certainty is just what medical image researchers demand. So the information within noisy area should be utilized to improve the prediction confidence.

Data insufficiency is another issue to apply supervised methods in medical images sometimes. It is difficult or even impossible to get noise-free or ground truth for some imaging modalities. Supervised methods may not work well under this condition. So we should leverage any available information to solve the data insufficiency. These two issues will be solved in the following parts.

\subsection{Content-Aware BMA Removal}
\label{subsec:cabr}
The uncertainty of context-guided inpainting models is the main drawback for the tasks demanding accurate results rather than plausible ones. We have observed that, even with heavy noise, some structure and texture information exist in BMA affected areas. To leverage such vessel clues, we introduce two information injection approaches,~\eg Gradient Statistics (GS) and Appearance Feature (AF). The proposed CABR can automatically learn reliable references from the injected information for BMA removal. The framework of CABR is illustrated in \cref{fig:framework}. 

\vspace{-5mm}
\paragraph{Gradient Statistics (GS).}
BMA usually causes high-intensity horizontal stripes in OCTA images. We observed that the horizontal gradients are hardly affected by BMA so we introduce the horizontal gradients as the structural information for CABR. Specifically, we apply the vertical Sobel operator in BMA areas to extract GS. The absolute value of response is injected into the model to capture more reference for BMA removal. Sobel operator is linear and efficient so only marginal computational cost is involved. 

\vspace{-5mm}
\paragraph{Appearance Feature (AF).}
GS provides a good reference for branch vessels but may not discern capillaries at relatively low intensities. Inspired by the deep denoising models, we define the BMA-affected areas as AF and keep AF, even noisy ones, as the additional input. With sufficient training samples, the BMA removal model can automatically learn the texture information to capture better vessel connectivities. The generation of training samples is discussed in \cref{subsec:selfsup}.

\vspace{-5mm}
\paragraph{Model Design.}
To overcome the limitations of context-guided inpainting model, we propose the content-based CABR for BMA removal in OCTA. The framework is illustrated in \cref{fig:framework}. With the two sets of information injection, our model can learn from both the content within BMA and the context outside BMA. It is thus more comprehensive and robust. 

The imbalance of BMA sizes is another challenge for BMA removal. Most BMAs are thin stripes and are easy to remove while severe BMAs, on the contrary, are only small part of dataset but harder to remove. For previous context-guided inpainting models, the imbalanced distribution leads to performance gap between gentle and severe BMAs. These models often fail to fill in extra wide stripes due to lacking both references and hard training samples. 

Different from previously free-form inpainting frameworks~\cite{liu2018image,ren2019structureflow,yu2019free}, our CABR focuses on correcting the vasculature in the centerline areas of input image. Reflection padding is used for the stripe located near boundary. This schema reduces the learning cost and model complexity. Furthermore, we can select more hard cases to train CABR so that it  works better when dealing with these broadly missing areas. 

\vspace{-5mm}
\paragraph{Model Architecture.}
CABR has an encoder-decoder CNN backbone, which stacks gated convolution layers (GatedConv)~\cite{yu2019free} to handle the stripe with various width in a noisy OCTA. As illustrated in \cref{fig:framework}, the backbone consists of 10 GatedConv layers~\cite{yu2019free}, 1 GatedConv layer with dilation~\cite{yu2015multi}, 2 DeConv layers and 1 regular Conv layer with \textit{sigmoid} activation function for prediction. GatedConv is a kind of attention convolution that learns a dynamic feature gate for each channel and each spatial location. We choose GatedConv due to its SOTA performance in inpainting. In addition, it saves around $50\%$ trainable parameters than regular convolution with the same channel number. Considering the relatively small dataset, such light-weight module is preferred to reduce overfitting. Deconvolution used in CABR is an upsampling layer followed by a GatedConv, as illustrated in \cref{fig:deconv}.

\subsection{Self-supervised Training}\label{subsec:selfsup}
As we mentioned before, data insufficiency is one of the main challenges in OCTA. Here, we train our model in a self-supervised way to resolve problem. We make use of the majority BMA-free areas to acquire acceptable training masks at low cost and generate as many as noisy images for training. With these easily collected masks and synthetic noisy images, we can train the BMA removal model in a self-supervised manner.

\vspace{-5mm}
\paragraph{Ground Truth Mask.}
The ground truth of vasculature mask for \textit{in~vivo} OCTA is either impossible or very difficult to acquire. Since OOF~\cite{law2010oriented} can produce relatively accurate vessel mask for the most area ($96.4\%$) without BMA, we resort to this defective mask with minimally additional manual annotation as the ground truth. To be specific, we only manually annotate the remaining $3.6\%$ BMA areas and combine defective mask in BMA-free areas to get the vasculature mask. This approach can produce high-recall masks at low-cost. The manually annotated part is only involved in testing and is not used in training.  All these masks are verified by experts.

\begin{figure}[!t]
     \centering
     \begin{subfigure}[b]{1.0\linewidth}
         \centering
         \includegraphics[width=\linewidth]{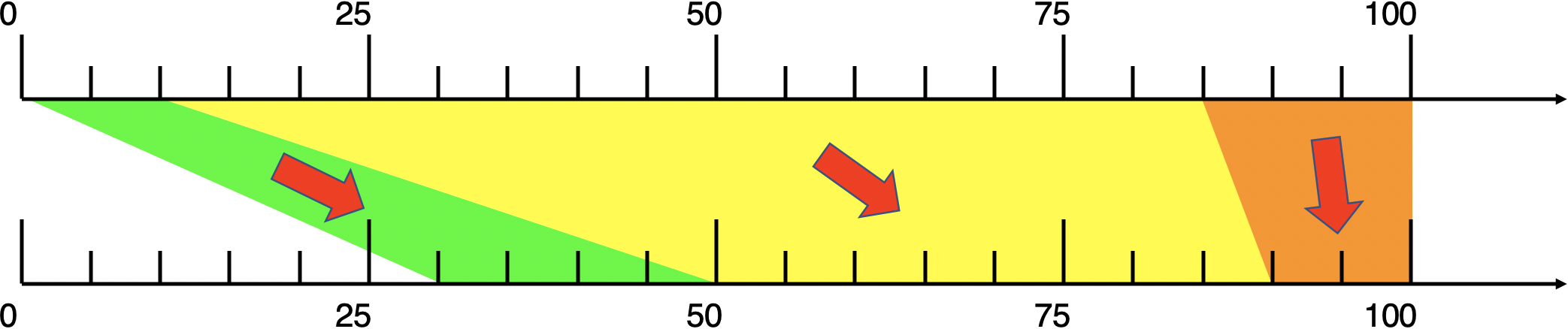}
         \caption{The schema of intensity mapping used in AdBMA.}
          \label{fig:synproj}
     \end{subfigure}
     \vfill
     \vspace{2.mm}
     \begin{subfigure}[b]{1.0\linewidth}
         \centering
         \includegraphics[width=\linewidth]{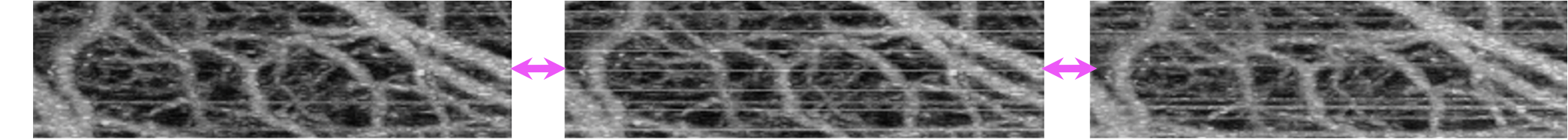}
         \caption{OCTA images.}
         \label{fig:synvis}
     \end{subfigure}
     \vfill
     \vspace{2.mm}
     \begin{subfigure}[b]{1.0\linewidth}
         \centering
         \includegraphics[width=\linewidth]{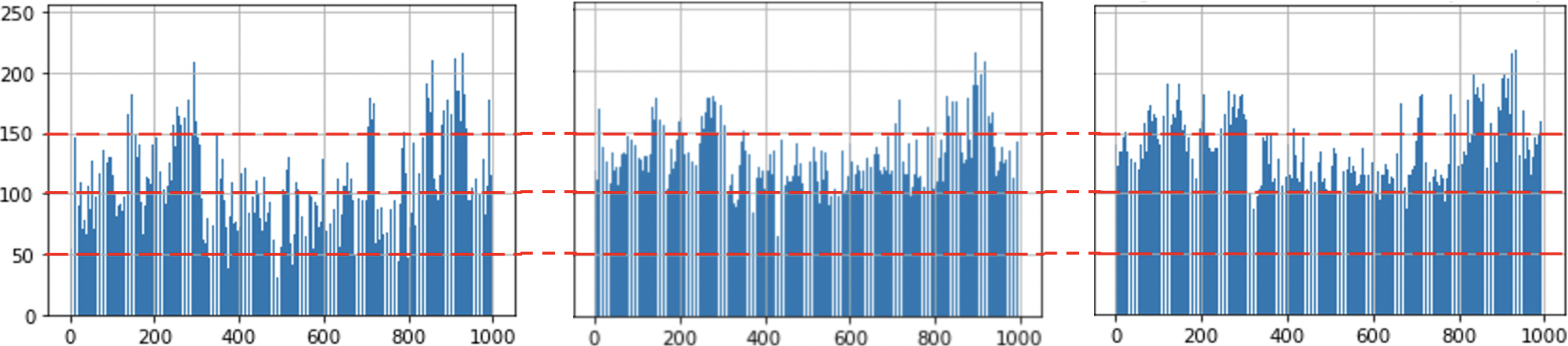}
         \caption{Intensity values of the row marked by double arrows in (b).}
         \label{fig:synbar}
     \end{subfigure}
     \caption{In (b) (c), from left to right: clear images, synthetic images and images affected by BMA.}
     \label{fig:synplot}
\end{figure}

\vspace{-5mm}
\paragraph{BMA Synthesis.}
We propose the Adaptive BMA Synthesis (AdBMA) approach to generate noisy images from BMA-free areas. After inspecting the data distributions, we find that BMA has different effects on high and low intensity pixels. In other words, BMA mainly uplifts the low-intensity pixels while keeping the maximums unchanged. So we employ a remapping schema to mimic the BMA influence on clear images. Given a clear image $\boldsymbol{I}$ and two sets of hyperparameters $(\mathcal{P}_{\text{low}},\mathcal{P}_{\text{high}})$ and  $(\mathcal{P}_{\text{base}},\mathcal{P}_{\text{low}^\prime},\mathcal{P}_{\text{high}^\prime})$, we synthesize BMA in four steps: (1) randomly map the intensities below $\mathcal{P}_{\text{low}}$ into $[\mathcal{P}_{\text{base}},\mathcal{P}_{\text{low}^\prime})$ (half-open interval); (2) map the intensities between $\mathcal{P}_{\text{low}}$ and $\mathcal{P}_{\text{high}}$ into $[\mathcal{P}_{\text{low}^\prime},\mathcal{P}_{\text{high}^\prime})$; (3) map the intensities above $\mathcal{P}_{\text{high}}$ into $[\mathcal{P}_{\text{high}^\prime},\max(\boldsymbol{I}))$; and (4) clip intensities within image maximum (usually 255) to avoid overflow. Gaussian noise is added in steps (2) and (3) to ensure uncertainty. The remapping schema is illustrated in \cref{fig:synproj}.

The synthesized noisy images have similar distribution with the real BMA, but more importantly they have corresponding ground truth masks. The benefits of synthesizing BMA are twofold: (1) we can generate as many plausible noisy images as needed; and (2) it fits our framework and enables learning from noisy data. 

\cref{fig:synvis} shows a BMA-free OCTA image, an image with synthetic BMA and an image with real BMA. Moreover, \cref{fig:synbar} is the bar plot of intensity values for the row marked by double arrows in \cref{fig:synvis}.  We can see that the synthetic OCTA image has not only plausible appearance but also similar intensity distribution. Thus the distribution gap between training and testing data is minimized.

\vspace{-5mm}
\paragraph{Training Scheme.}
The train/val splitting is area-based, which is different from the typical supervised approach. Specifically, every image is split into the BMA-free (clear) area for training and the BMA-affected (noisy) area for evaluation. During training, only the loss from the clear area with synthetic BMA is used. During testing, only the Dice score in the BMA-affected area is measured. Thus the model works in a \textit{self-supervised} way and can utilize as many training samples as possible.

\section{Experiments}
\label{sec:exp}
We evaluate the proposed BMA removal framework on a mouse cortex OCTA dataset. The results are compared with the state-of-the-art motion correction model~\cite{li2021deep} and a GatedConv-based inpainting model~\cite{yu2019free}. We re-implement the GatedConv-based model for inpainting the stripe region.

\subsection{Dataset}
\label{subsec:dataset}
We collect 39 OCTA images of mouse cortex with vessel masks. The noise level, defined as $(\text{BMA-affected area}/\text{Total area})$, varies throughout the dataset. Note that, not all images have BMA noise. We validate on the BMA-affected areas from 14 OCTAs. According to the noise level, OCTAs in the validation set are further classified as the easy set ($\text{noise}\!\!<\!\!2\%$, 4 OCTAs), medium set ($2\!\!\le\!\!\text{noise}\!\!<\!\!4\%$, 7 OCTAs) and hard set ($\text{noise}\!\!\ge\!\!4\%$, 3 OCTAs). We further collect 6 samples to extend the validation set. The BMA removal performance is measured by the Sørensen–Dice coefficient (Dice) in the rest of the paper, unless otherwise specified.



\vspace{-5mm}
\paragraph{Training Details.}
The hyperparameters for BMA synthesis are $(\mathcal{P}_{\text{low}}=\mathcal{P}_{10\%},\mathcal{P}_{\text{high}}=\mathcal{P}_{85\%})$ and  $(\mathcal{P}_{\text{base}}=\mathcal{P}_{30\%},\mathcal{P}_{\text{low}^\prime}=\mathcal{P}_{50\%},\mathcal{P}_{\text{high}^\prime}=\mathcal{P}_{90\%})$ where $\mathcal{P}_{k\%}$ is the $k$-th percentile of input image $\boldsymbol{I}$. As for the re-implemented GatedConv-based inpainting model, we randomly select $2\!\!\sim\!\!20\%$ rows within each patch for training. As for CABR, we select center rows in each patch with width randomly from $1$ to $11$ for training. Furthermore, we use data augmentation, such as random cropping, horizontal and vertical flipping, on the training set.

\vspace{-5mm}
\paragraph{Implementation.}
The CNN backbone used for all experiments is illustrated in \cref{fig:framework}. All convolutional layers use $3\times3$ kernels. The downsampling rate, stride rate and dilation rate are set to 2. The backbone has 16 feature maps in the first level, which increases up to 64 when the level gets deeper. All models are trained with Dice loss and Adam optimizer for 1,200 epochs. The learning rate starts from $10^{-4}$ and halves if no improvement is observed for 20 epochs. We used $256\times496$ input size and batch size 12 for GatedConv training. We used $64\times496$ input size and batch size 48 for CABR training. All models are trained on a single NVIDIA GeForce GTX Titan Xp GPU.

\subsection{BMA Removal Results}
\begin{table}[!t]
  \centering
  \caption{Performance on the mouse cortex OCTA dataset. GatedConv* is re-implemented and trained on the OCTA dataset. ``All*" is the extended validation set with ``All" and 6 extra samples.}
\begin{tabular}
    { @{\hspace{0.5mm}}c@{\hspace{1.2mm}} @{\hspace{0.5mm}}c@{\hspace{1.2mm}} @{\hspace{0.5mm}}c@{\hspace{.2mm}}  @{\hspace{1.5mm}}c@{\hspace{1.2mm}} @{\hspace{1.5mm}}c@{\hspace{1.2mm}}  @{\hspace{1.5mm}}c@{\hspace{1.2mm}} }
    \toprule
    Method &  Easy  & Medium & Hard  & All & All*\\
    \midrule
    \small{OOF~\cite{law2010oriented}} & 45.76 & 42.70 & 43.72 & 43.79 & 42.93 \\
    \small{Li~\etal~\cite{li2021deep}} & 68.83 & 68.51 & 60.70 & 66.93 & 70.21 \\
    \small{GatedConv*~\cite{yu2019free}} & \multicolumn{1}{r}{77.41} & 73.48 & 74.37 & 74.80 & 77.03\\
    \midrule
    \small{CABR}\tiny{\textit{(SynBMA)}} & 84.99 & 81.98 & 79.22 & 82.24 & 83.92 \\
    \small{CABR}\tiny{\textit{(SynBMA,GE)}} & \textbf{85.71} & \textbf{82.67} & \textbf{79.69} & \textbf{82.90} & \textbf{84.54} \\
    \bottomrule
    \end{tabular}%
  \label{tab:exp_all}%
\end{table}%

\begin{figure}[!t]
    \centering
    \includegraphics[width=0.98\linewidth]{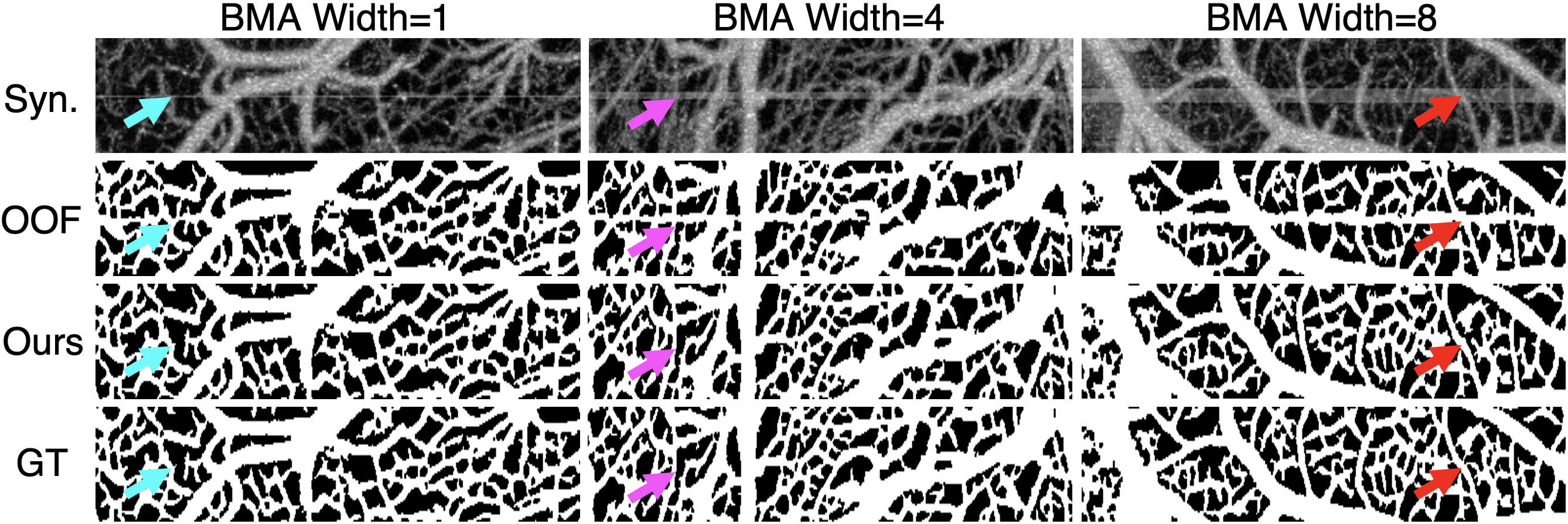}
    \caption{Denoising result for synthesized BMA of width \{1,4,8\}.}
    \label{fig:synall}
\end{figure}

\begin{figure*}
     \centering
     \begin{subfigure}[b]{0.90\textwidth}
         \centering
         \includegraphics[width=\textwidth]{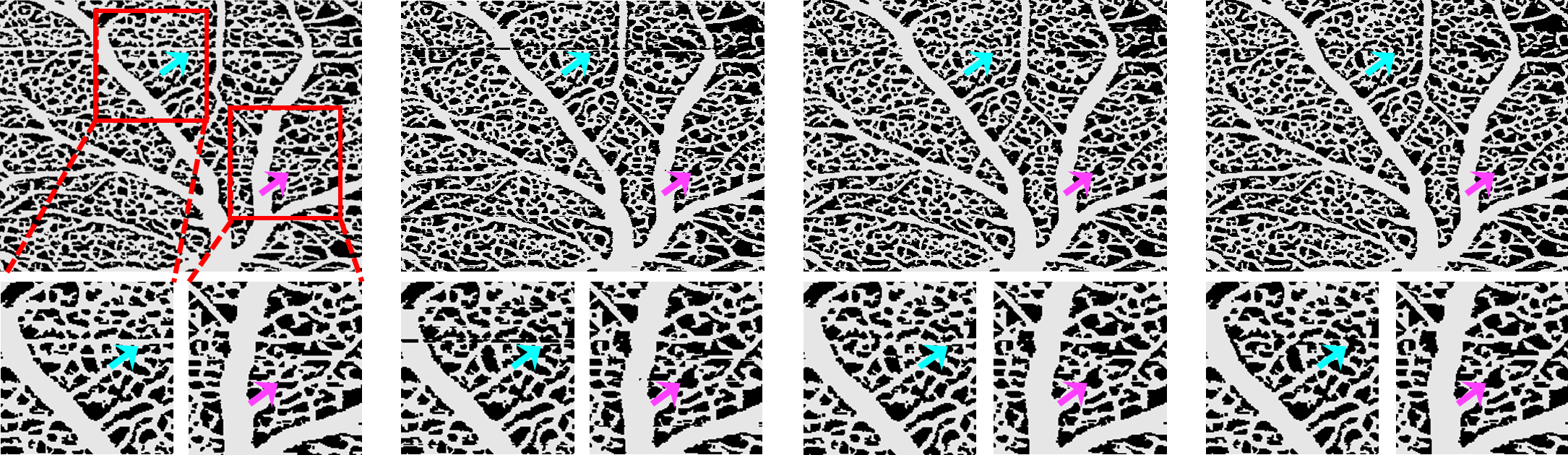}
         \label{fig:example1_mask}
     \end{subfigure}
     \vfill
     \begin{subfigure}[b]{0.90\textwidth}
         \centering
         \includegraphics[width=\textwidth]{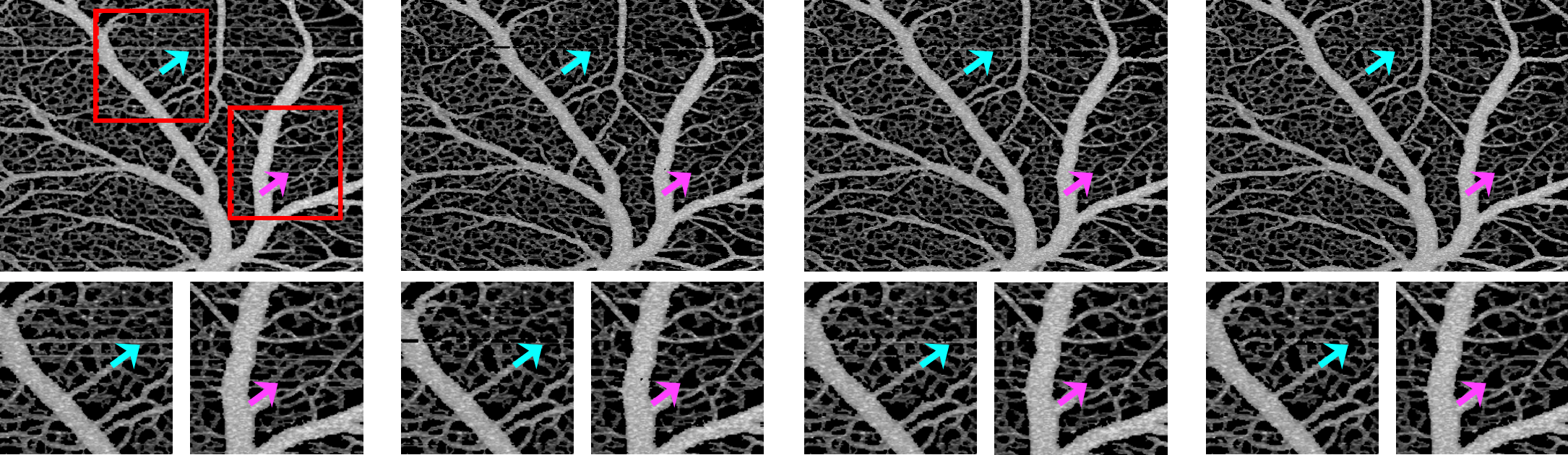}
         \label{fig:example1}
     \end{subfigure}
     \vfill
     \begin{subfigure}[b]{0.90\textwidth}
         \centering
         \includegraphics[width=\textwidth]{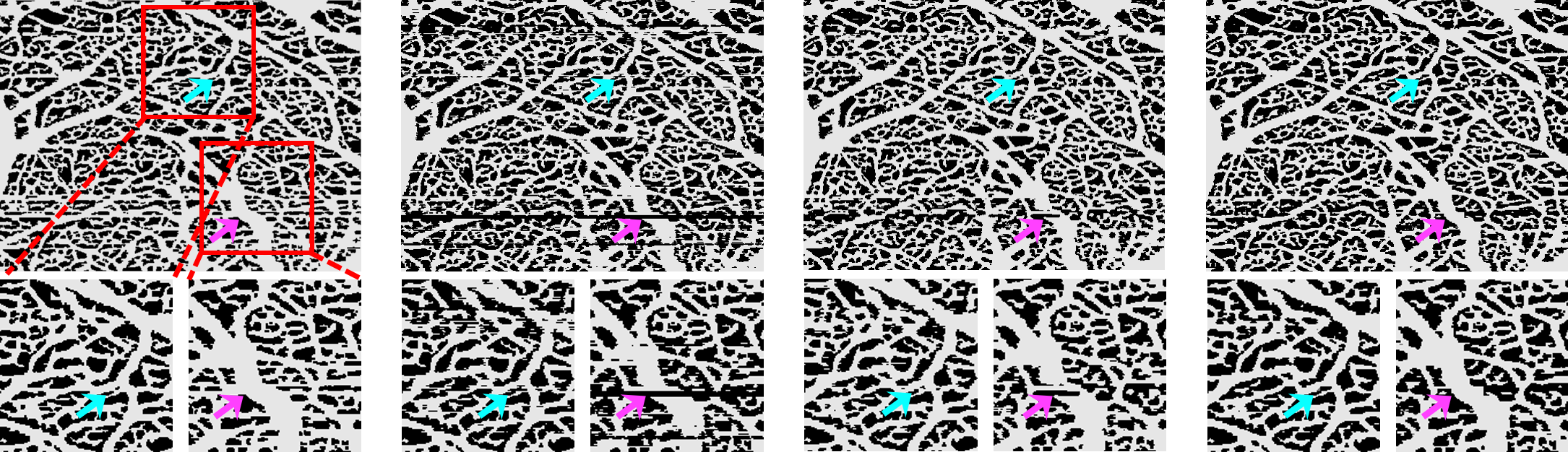}
         \label{fig:example2_mask}
     \end{subfigure}
     \vfill
     \begin{subfigure}[b]{0.90\textwidth}
         \centering
         \includegraphics[width=\textwidth]{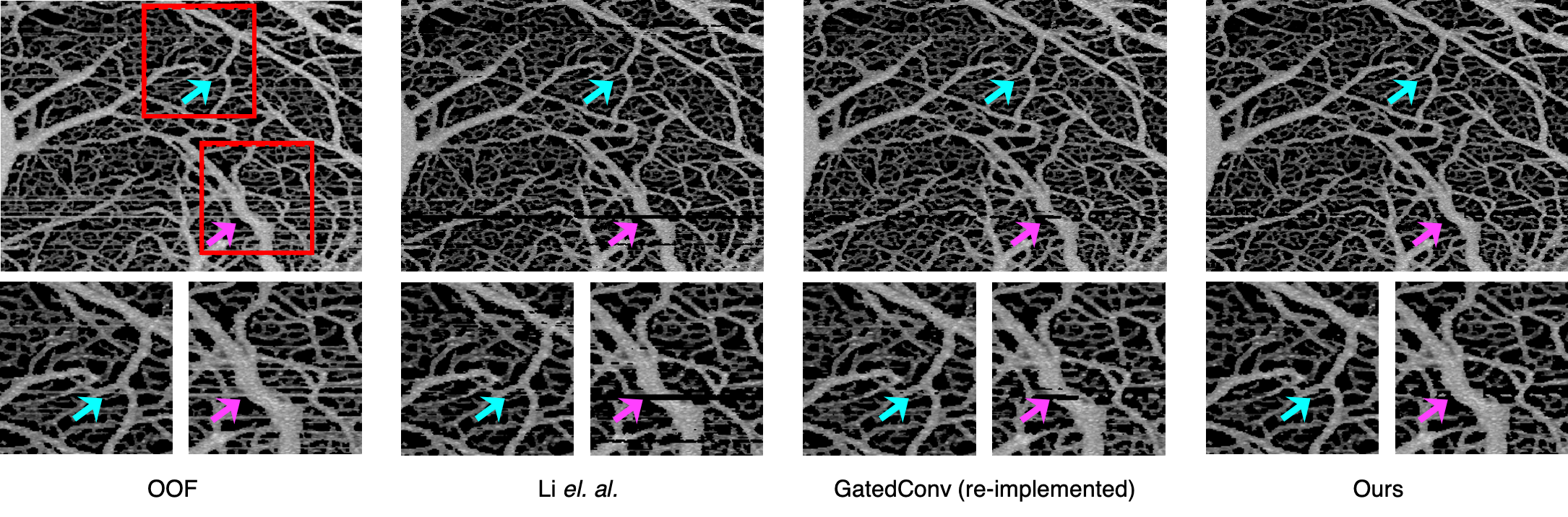}
         \label{fig:example2}
     \end{subfigure}
        \caption{Two sets of OCTA masks and enhanced images of different methods. The first and third rows are the masks of OOF~\cite{law2010oriented}, Li's method~\cite{li2021deep}, our re-implemented GatedConv-based inpainting model~\cite{yu2019free} and the proposed CABR, respectively. The corresponding enhanced images are in the second and fourth rows. The zoom-in results within the rectangles in dashed line are also illustrated for better comparison. The cyan and pink arrows highlight more details in the zoom-in panels.}
        \label{fig:example}
\end{figure*}

\begin{figure}[t]
  \centering
   \includegraphics[width=1.0\linewidth]{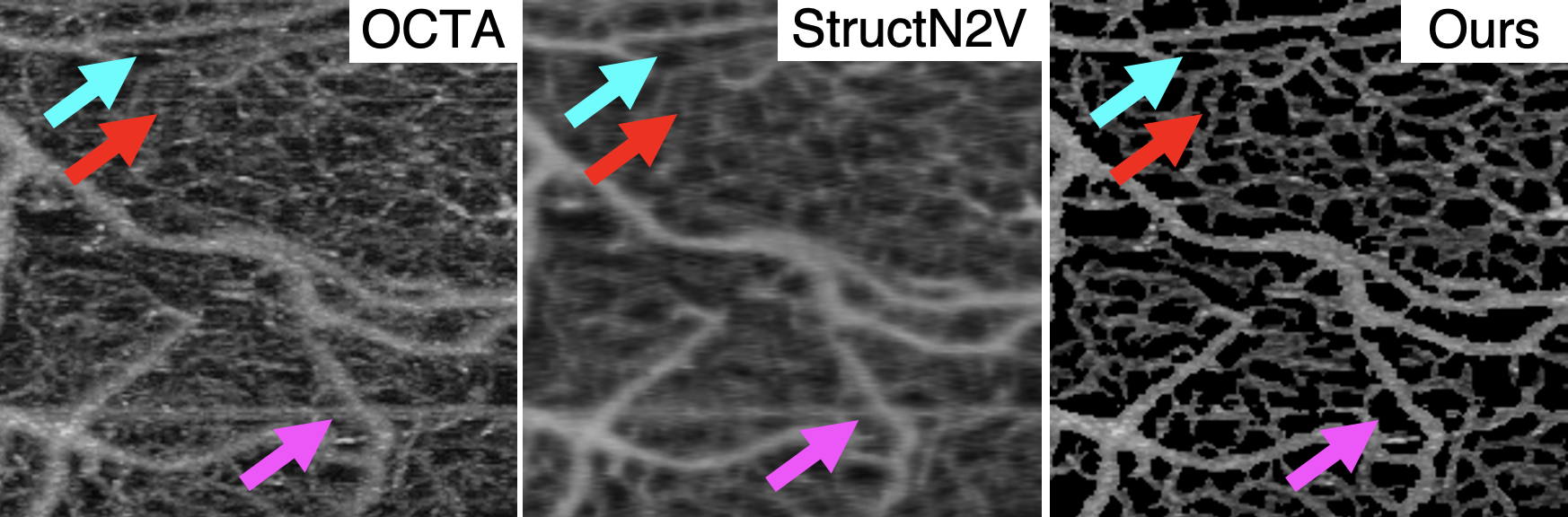}
   \caption{The denoising result of StructN2V~\cite{broaddus2020removing} and our method.}
   \label{fig:structn2v}
\end{figure}

To evaluate the performance of the proposed CABR framework, we compare it with the state-of-the-art OCTA motion correction model by Li~\etal~\cite{li2021deep}. The result of OOF~\cite{law2010oriented} is the baseline. To the best of our knowledge, there are no other publications that pertain directly to BMA removal in OCTA. So we also re-implement a GatedConv-based inpainting model~\cite{yu2019free}, for inpainting the BMA-affected region. The re-implemented inpainting model has a similar amount of parameters as CABR and is trained with the same setting.


\vspace{-5mm}
\paragraph{Quantitative Result.}
The Dice scores on the mouse cortex OCTA dataset of different methods are in \cref{tab:exp_all}. The result shows a clear improvement of our method over the OOF baseline. Besides, our approach significantly outperforms the state-of-the-art model~\cite{li2021deep} by $15.97\%$ and $14.33\%$ in the validation set and the extended validation set, respectively. The proposed CABR also achieves the best performance in all subsets, which evidences the effectiveness and advantage of content-based learning. Even in the hard subset with larger and more dense BMA across OCTA, our CABR surpasses the re-implemented GatedConv by $5.32\%$, suggesting that CABR can learn more connectivities from noise input for BMA removal. The experiment results also show that adding GS further improves the performance of CABR.

\vspace{-5mm}
\paragraph{Qualitative Result.}
The BMA removal results for the synthesized BMA of width 1, 4 and 8 are illustrated in \cref{fig:synall}. It shows that our method can handle severe BMA and recover vasculature while the baseline~\cite{law2010oriented} fails.

The experiment results for two validation samples are illustrated in the first and third rows of \cref{fig:example}. We compare our method with OOF~\cite{law2010oriented} (only for reference, no BMA removal effect), Li~\etal's method~\cite{li2021deep}, our re-implemented GatedConv-based inpainting model~\cite{yu2019free}. We can see from the first and third rows of \cref{fig:example} that our method removes most of the false positive masks caused by BMA while achieving the best vasculature connectivity at the same time. Li~\etal's method~\cite{li2021deep} can also remove some BMA but the low recall is the main drawback. Specifically, thin BMA stripes can be restored while thicker ones can not. On the contrary, the re-implemented GatedConv shows better connectivity than Li~\etal's method~\cite{li2021deep} but some BMAs are remaining. The cyan and pink arrows highlight more details in the zoom-in panels of \cref{fig:example}.

The second OCTA is harder to handle because the stripes are relatively broader and more densely distributed. From the third and fourth rows of \cref{fig:example}, we can see that Li~\etal's method~\cite{li2021deep} can not fill the gap in the second zoom-in area and the GatedConv-based inpainting model has more artifacts. Our method produces a complete mask with better connectivity and fewer artifacts. 

We also compare with the self-supervised StructN2V~\cite{broaddus2020removing} and the results in \cref{fig:structn2v} suggest that StructN2V is not suitable for our task:  incorrect structure removal (cyan/red arrow), suffering from wide BMA (pink arrow), \etc. 


\vspace{-5mm}
\paragraph{OCTA Enhancement.}
OCTA enhancement based on vesselness mask is a regular procedure for biomedical analysis. Usually, this procedure highlights vasculature while suppressing background noise, and thus improves contrast. The enhanced OCTA is generated by multiplying the raw OCTA with corresponding vessel mask. The second and fourth rows of \cref{fig:example} illustrate the enhancement results of different approaches on two sets of BMA-affected OCTAs. Our method removes most of the BMA artifacts and produces an OCTA with stronger contrast, as illustrated in the zoom-in panels as the dashed rectangle areas. From the second OCTA, where BMA in the bottom area degrades image quality severely, we can see that our method still removes most of the BMAs while preserving vessel structures, thus showcasing the effectiveness of our proposed method.

\begin{table}[t]
  \centering
  \caption{Ablation study of CABR on the mouse cortex OCTA dataset. ``Abs" means the absolute value. ``Abs-BMA" means absolute GS only in BMA. ``Conv" means the regular convolution.} 
  \begin{tabular}
    {@{\hspace{1.mm}}c@{\hspace{1.mm}} @{\hspace{1.mm}}c@{\hspace{1.mm}}| c@{\hspace{1.mm}}   @{\hspace{1.mm}}c@{\hspace{1.mm}} @{\hspace{1.mm}}c@{\hspace{1.mm}}| c@{\hspace{1.mm}}   @{\hspace{1.mm}}c@{\hspace{1.mm}}| c@{}}
    \toprule
    \multicolumn{2}{c|}{\small{Appearance}} & \multicolumn{3}{c|}{\small{Gradient Statistics}} & \multicolumn{2}{c|}{\small{Module}} & \small{Dice$\uparrow$\footnotesize{($\%$)}} \\
    \footnotesize{Gauss} & \footnotesize{AdBMA} & \footnotesize{Naive} & \footnotesize{Abs}  & \footnotesize{Abs-BMA} & \footnotesize{Conv} & \footnotesize{Gated} &\\
    \midrule
    \checkmark    &       &       &       &       &  & \checkmark & 81.29 \\
    \hline& \checkmark    &       &       &       &  & \checkmark & 82.24 \\
    \hline  & \checkmark    & \checkmark    &       &       &  & \checkmark & 80.08 \\
    \hline      & \checkmark    &       & \checkmark    &       &  & \checkmark & 82.38 \\
    \hline      & \checkmark    &       & \checkmark    &       & \checkmark &  & 82.57 \\
    \hline  & \checkmark    &       &       & \checkmark    &  & \checkmark & \textbf{82.90} \\
    \bottomrule
  \end{tabular}%
  \label{tab:ablation}%
\end{table}%

\subsection{Ablation Study}
To study the effects of each module, we conduct the ablation experiments on the mouse cortex OCTA dataset. All experiments are based on the same training setting in \cref{subsec:dataset}. For AF injection, we compare the proposed AdBMA with a naive implementation that adds Gaussian noise. Both methods generate high-intensity stripes like BMA artifacts. However, the Gaussian method does not have a blurring effect so the result is less plausible. For GS injection, the \textit{naive} stands for the gradient of Sobel operator. \textit{Abs} and \textit{Abs-BMA} stand for the absolute value of Sobel results on the whole image and only the BMA area, respectively. To report the contribution of GatedConv module, we also conduct an ablative study and find that GatedConv does outperform regular convolution by $0.33\%$ with only about $50\%$ trainable parameters.

\vspace{-5mm}
\paragraph{Ablation of Information Injection.}
As shown in \cref{tab:ablation}, we first investigate the effectiveness of AF injection with synthetic BMA. ``Gauss" means adding Gaussian noise directly to generate training samples. Our proposed AdBMA outperforms the Gaussian-based synthetic approach by $1.66\%$, which verifies that (1) texture information within BMA, even noisy, still benefits the recovery of vasculature; and (2) the blurring effect from AdBMA improves the performance. We also evaluate the effectiveness of GS as shown in \cref{tab:ablation}. The performance of CABR is improved by $2.30\%$ for Abs and $2.82\%$ for Abs-BMA, respectively, which states the validity of GS in BMA removal. During experiments, we find that GS leads to faster convergence. CABR benefits less from GS than the synthetic BMA feature. One possible reason is that convolution layers may have learned similar or even more powerful kernels than the Sobel operator.

\vspace{-5mm}
\paragraph{Ablation of Model Size.}
With the same modules in the CNN backbone, we investigate how the model size influences the performance of BMA removal. \cref{tab:channel} lists the result of CABR with different sizes. A larger channel number means more parameters and a larger model size. The 8-channel model only achieves $78.70\%$ Dice coefficient, which is less than 16-channel and 32-channel models. The results demonstrate that 8-channel is insufficient to capture all vessel information. Also, a larger model size, \ie, 32-channel in the first layer, does not boost the performance. The 16-channel model outperforms the 32-channel by $0.46\%$ with about $75\%$ less trainable parameters, suggesting that the OCTA dataset may not support training the model with over 140K parameters.

\begin{table}
  \centering
  \caption{Ablation study of CABR for different model sizes on the mouse cortex OCTA dataset. \# Channels is the channel number for the first layer of CNN backbone. \# Parameters is the number of trainable parameters for the whole model.}
    \begin{tabular}{ccc}
    \toprule
    \multicolumn{1}{l}{\# Channels} & \# Parameters & \multicolumn{1}{l}{Dice$\uparrow$\small{($\%$)}} \\
    \midrule
    8     & 33.7K      & 78.70 \\
    16    & 133.9K   & \textbf{82.90} \\
    32    & 534.0K   & 82.44 \\
    \bottomrule
    \end{tabular}%
  \label{tab:channel}%
\end{table}%


\section{Conclusion}
\label{sec:con}
In this paper, we propose a self-supervised content-aware model to remove BMA and improve image quality in OCTA. First, the GS-based structural information and AF are extracted from the BMA area and injected into the model to capture more vasculature connectivity. Second, we train the model in a self-supervised manner with easily collected defective masks. With the rich structural and appearance information carried in the BMA stripe region as references, our model can remove BMAs and produce a better vasculature mask. Compared with other context-guided inpainting models, our content-aware model can learn guidance directly from noisy images and thus is more robust. Experiments on a mouse cortex OCTA dataset demonstrate that our model can remove most BMAs, even the huge and inconsistent ones, while existing methods fail. The current BMA synthesis approach is manually designed and still has room for improvement. Incorporating domain knowledge of BMA formulation or introducing a generative model for synthesis are potential research directions.

\vspace{-2.5mm}
\paragraph{Acknowledgment.} This work was supported in part by NSF Grants 1814745 and 2006665, and NIH grants R01DA029718 and RF1DA048808.


{\small
\bibliographystyle{ieee_fullname}
\bibliography{egbib_ab}

\begin{thebibliography}{10}\itemsep=-1pt

\bibitem{broaddus2020removing}
Coleman Broaddus, Alexander Krull, Martin Weigert, Uwe Schmidt, and Gene Myers.
\newblock Removing structured noise with self-supervised blind-spot networks.
\newblock In {\em ISBI}, 2020.

\bibitem{camino2016evaluation}
Acner Camino, Miao Zhang, Simon~S Gao, Thomas~S Hwang, Utkarsh Sharma, David~J
  Wilson, David Huang, and Yali Jia.
\newblock Evaluation of artifact reduction in optical coherence tomography
  angiography with real-time tracking and motion correction technology.
\newblock {\em Biomed. Opt. Express}, 7(10):3905--3915, 2016.

\bibitem{chen2016trainable}
Yunjin Chen and Thomas Pock.
\newblock Trainable nonlinear reaction diffusion: A flexible framework for fast
  and effective image restoration.
\newblock {\em TPAMI}, 39(6):1256--1272, 2016.

\bibitem{dabov2007image}
Kostadin Dabov, Alessandro Foi, Vladimir Katkovnik, and Karen Egiazarian.
\newblock Image denoising by sparse 3-d transform-domain collaborative
  filtering.
\newblock {\em TIP}, 16(8):2080--2095, 2007.

\bibitem{devalla2019deep}
Sripad~Krishna Devalla, Giridhar Subramanian, Tan~Hung Pham, Xiaofei Wang,
  Shamira Perera, Tin~A Tun, Tin Aung, Leopold Schmetterer, Alexandre~H Thiery,
  and Micha{\"e}l~JA Girard.
\newblock A deep learning approach to denoise optical coherence tomography
  images of the optic nerve head.
\newblock {\em Sci. Rep.}, 9(1):1--13, 2019.

\bibitem{fadnavis2020patch2self}
Shreyas Fadnavis, Joshua Batson, and Eleftherios Garyfallidis.
\newblock Patch2self: denoising diffusion mri with self-supervised learning.
\newblock In {\em NeurIPS}, 2020.

\bibitem{frangi1998multiscale}
Alejandro~F Frangi, Wiro~J Niessen, Koen~L Vincken, and Max~A Viergever.
\newblock Multiscale vessel enhancement filtering.
\newblock In {\em MICCAI}, 1998.

\bibitem{huang1991optical}
David Huang, Eric~A Swanson, Charles~P Lin, Joel~S Schuman, William~G Stinson,
  Warren Chang, Michael~R Hee, Thomas Flotte, Kenton Gregory, Carmen~A
  Puliafito, et~al.
\newblock Optical coherence tomography.
\newblock {\em Science}, 254(5035):1178--1181, 1991.

\bibitem{jain2008natural}
Viren Jain and Sebastian Seung.
\newblock Natural image denoising with convolutional networks.
\newblock In {\em NeurIPS}, 2008.

\bibitem{krull2019noise2void}
Alexander Krull, Tim-Oliver Buchholz, and Florian Jug.
\newblock Noise2void-learning denoising from single noisy images.
\newblock In {\em CVPR}, 2019.

\bibitem{law2008three}
Max~WK Law and Albert~CS Chung.
\newblock Three dimensional curvilinear structure detection using optimally
  oriented flux.
\newblock In {\em ECCV}, 2008.

\bibitem{law2010oriented}
Max~WK Law and Albert~CS Chung.
\newblock An oriented flux symmetry based active contour model for three
  dimensional vessel segmentation.
\newblock In {\em ECCV}, 2010.

\bibitem{lehtinen2018noise2noise}
Jaakko Lehtinen, Jacob Munkberg, Jon Hasselgren, Samuli Laine, Tero Karras,
  Miika Aittala, and Timo Aila.
\newblock Noise2noise: Learning image restoration without clean data.
\newblock In {\em ICML}, 2018.

\bibitem{li2021deep}
Ang Li, Congwu Du, and Yingtian Pan.
\newblock Deep-learning-based motion correction in optical coherence tomography
  angiography.
\newblock {\em J. Biophotonics}, 2021.

\bibitem{li2017automated}
Ang Li, Jiang You, Congwu Du, and Yingtian Pan.
\newblock Automated segmentation and quantification of oct angiography for
  tracking angiogenesis progression.
\newblock {\em Biomed. Opt. Express}, 8(12):5604--5616, 2017.

\bibitem{li2018automated}
Ang Li, Guang Zeng, Congwu Du, Huiping Zhang, and Yingtian Pan.
\newblock Automated motion-artifact correction in an octa image using tensor
  voting approach.
\newblock {\em Appl. Phys. Lett.}, 113(10), 2018.

\bibitem{liu2018image}
Guilin Liu, Fitsum~A Reda, Kevin~J Shih, Ting-Chun Wang, Andrew Tao, and Bryan
  Catanzaro.
\newblock Image inpainting for irregular holes using partial convolutions.
\newblock In {\em ECCV}, 2018.

\bibitem{medioni2000tensor}
G{\'e}rard Medioni, Chi-Keung Tang, and Mi-Suen Lee.
\newblock Tensor voting: Theory and applications.
\newblock In {\em RFIA}, 2000.

\bibitem{querques2017optical}
Lea Querques, Chiara Giuffr{\`e}, Federico Corvi, Ilaria Zucchiatti, Adriano
  Carnevali, Luigi~A De~Vitis, Giuseppe Querques, and Francesco Bandello.
\newblock Optical coherence tomography angiography of myopic choroidal
  neovascularisation.
\newblock {\em Br. J. Ophthalmol.}, 101(5):609--615, 2017.

\bibitem{ren2019structureflow}
Yurui Ren, Xiaoming Yu, Ruonan Zhang, Thomas~H Li, Shan Liu, and Ge Li.
\newblock Structureflow: Image inpainting via structure-aware appearance flow.
\newblock In {\em ICCV}, 2019.

\bibitem{samara2017quantification}
Wasim~A Samara, Abtin Shahlaee, Murtaza~K Adam, M~Ali Khan, Allen Chiang,
  Joseph~I Maguire, Jason Hsu, and Allen~C Ho.
\newblock Quantification of diabetic macular ischemia using optical coherence
  tomography angiography and its relationship with visual acuity.
\newblock {\em Ophthalmology}, 124(2):235--244, 2017.

\bibitem{vakoc2009three}
Benjamin~J Vakoc, Ryan~M Lanning, James~A Tyrrell, Timothy~P Padera, Lisa~A
  Bartlett, Triantafyllos Stylianopoulos, Lance~L Munn, Guillermo~J Tearney,
  Dai Fukumura, Rakesh~K Jain, et~al.
\newblock Three-dimensional microscopy of the tumor microenvironment in vivo
  using optical frequency domain imaging.
\newblock {\em Nat. Med.}, 15(10):1219--1223, 2009.

\bibitem{waheed2016optical}
Nadia~K Waheed, Eric~M Moult, James~G Fujimoto, and Philip~J Rosenfeld.
\newblock Optical coherence tomography angiography of dry age-related macular
  degeneration.
\newblock {\em Dev. Ophthalmol.}, 56:91--100, 2016.

\bibitem{wei2017automatic}
David~Wei Wei, Anthony~J Deegan, and Ruikang~K Wang.
\newblock Automatic motion correction for in vivo human skin optical coherence
  tomography angiography through combined rigid and nonrigid registration.
\newblock {\em J. Biomed. Opt.}, 22(6), 2017.

\bibitem{weigert2018content}
Martin Weigert, Uwe Schmidt, Tobias Boothe, Andreas M{\"u}ller, Alexandr
  Dibrov, Akanksha Jain, Benjamin Wilhelm, Deborah Schmidt, Coleman Broaddus,
  Si{\^a}n Culley, et~al.
\newblock Content-aware image restoration: pushing the limits of fluorescence
  microscopy.
\newblock {\em Nat. Methods}, 15(12):1090--1097, 2018.

\bibitem{wu2021contrastive}
Haiyan Wu, Yanyun Qu, Shaohui Lin, Jian Zhou, Ruizhi Qiao, Zhizhong Zhang, Yuan
  Xie, and Lizhuang Ma.
\newblock Contrastive learning for compact single image dehazing.
\newblock In {\em CVPR}, 2021.

\bibitem{yu2015multi}
Fisher Yu and Vladlen Koltun.
\newblock Multi-scale context aggregation by dilated convolutions.
\newblock In {\em ICLR}, 2016.

\bibitem{yu2019free}
Jiahui Yu, Zhe Lin, Jimei Yang, Xiaohui Shen, Xin Lu, and Thomas~S Huang.
\newblock Free-form image inpainting with gated convolution.
\newblock In {\em ICCV}, 2019.

\bibitem{zhang2018ffdnet}
Kai Zhang, Wangmeng Zuo, and Lei Zhang.
\newblock Ffdnet: Toward a fast and flexible solution for cnn-based image
  denoising.
\newblock {\em TIP}, 27(9):4608--4622, 2018.

\bibitem{zhou2020awgn}
Yuqian Zhou, Jianbo Jiao, Haibin Huang, Yang Wang, Jue Wang, Honghui Shi, and
  Thomas Huang.
\newblock When awgn-based denoiser meets real noises.
\newblock In {\em AAAI}, 2020.

\end{thebibliography}
}

\end{document}